\documentclass[
]{ceurart}

\usepackage{listings}
\usepackage{graphicx}
\usepackage{booktabs}
\usepackage{multirow}
\usepackage{hyperref}
\begin{document}

\copyrightyear{2026}
\copyrightclause{Copyright for this paper by its authors.
  Use permitted under Creative Commons License Attribution 4.0
  International (CC BY 4.0).}
\conference{DMKG'26: 2nd International Workshop on Data Management for Knowledge Graphs, October 2026, Bari, Italy}
\title{Cost Characterization of Vertically Partitioned Federated Knowledge Graphs}

\author{Md Saikat Islam Khan Bappy}[%
  orcid=0009-0009-1768-6102,
  email=islamm9@rpi.edu
]

\author{Oshani Seneviratne}[%
  orcid=0000-0001-8518-917X,
  email=senevo@rpi.edu
]

\address{Rensselaer Polytechnic Institute,
Troy, NY 12180, USA}
\begin{abstract}
Knowledge graphs are increasingly distributed across autonomous organizations that share an entity space but own disjoint subsets of relations, forming a vertical partition. Answering a multi-hop query may require combining facts from several silos, making the partitioning strategy a key data management decision that affects communication, indexing, load balance, and query latency. However, the costs associated with different partitioning strategies remain insufficiently studied. We formalize vertical partitioning as a design space and compare four strategies: semantic domain grouping, frequency-balanced partitioning, co-occurrence graph-cut partitioning, and random partitioning. We evaluate them using five metrics: communication cost, candidate index size, cross-silo path length, load balance, and end-to-end query latency. Three of the five prove to be determined by the graph and the silo count rather than by the partition, which reduces the design problem to two conflicting axes, cross-silo path length and load balance. Experiments on MetaQA and PathQuestion use a fixed federated knowledge graph question-answering architecture based on TransE embeddings and a frozen BERT encoder across three silo configurations. By keeping the learning model unchanged, we isolate the effect of partitioning and show that the trade-off between locality and balance holds only where each silo can hold several relations, weakening as the number of silos increases. The study provides practical guidance for deployments constrained by cross-silo reasoning or by silo load.
\end{abstract}

\begin{keywords}
Federated Knowledge Graphs \sep
Vertical Partitioning \sep
Federated Question Answering \sep
Data Management Cost Analysis \sep
Graph Partitioning
\end{keywords}

\maketitle

\section{Introduction}
\label{sec:intro}

The Web of knowledge graphs is not a single queryable store but an
ecosystem of autonomous sources. In many real deployments, facts about the
same entities are distributed across organizations that each hold a
different slice of the relation vocabulary: a film studio records who
directed a film, a streaming platform who starred in it, a metadata service
its genre. All three describe the same entities, but no party holds the
whole graph, and raw data cannot be centralized due to governance,
commercial sensitivity, and data sovereignty constraints. This is a
\emph{vertical partition} of a knowledge graph: the entity space is shared,
while the relations are split into disjoint, privately held subsets. It
differs fundamentally from the horizontal federation studied by most
federated knowledge graph work~\cite{hu2025learning, gunti2025federated,
chen2024unaligned}, where every party holds the same relations over
different entities and can often answer a query within its own shard.

\begin{figure}[t]
  \centering
  \includegraphics[width=0.60\linewidth]{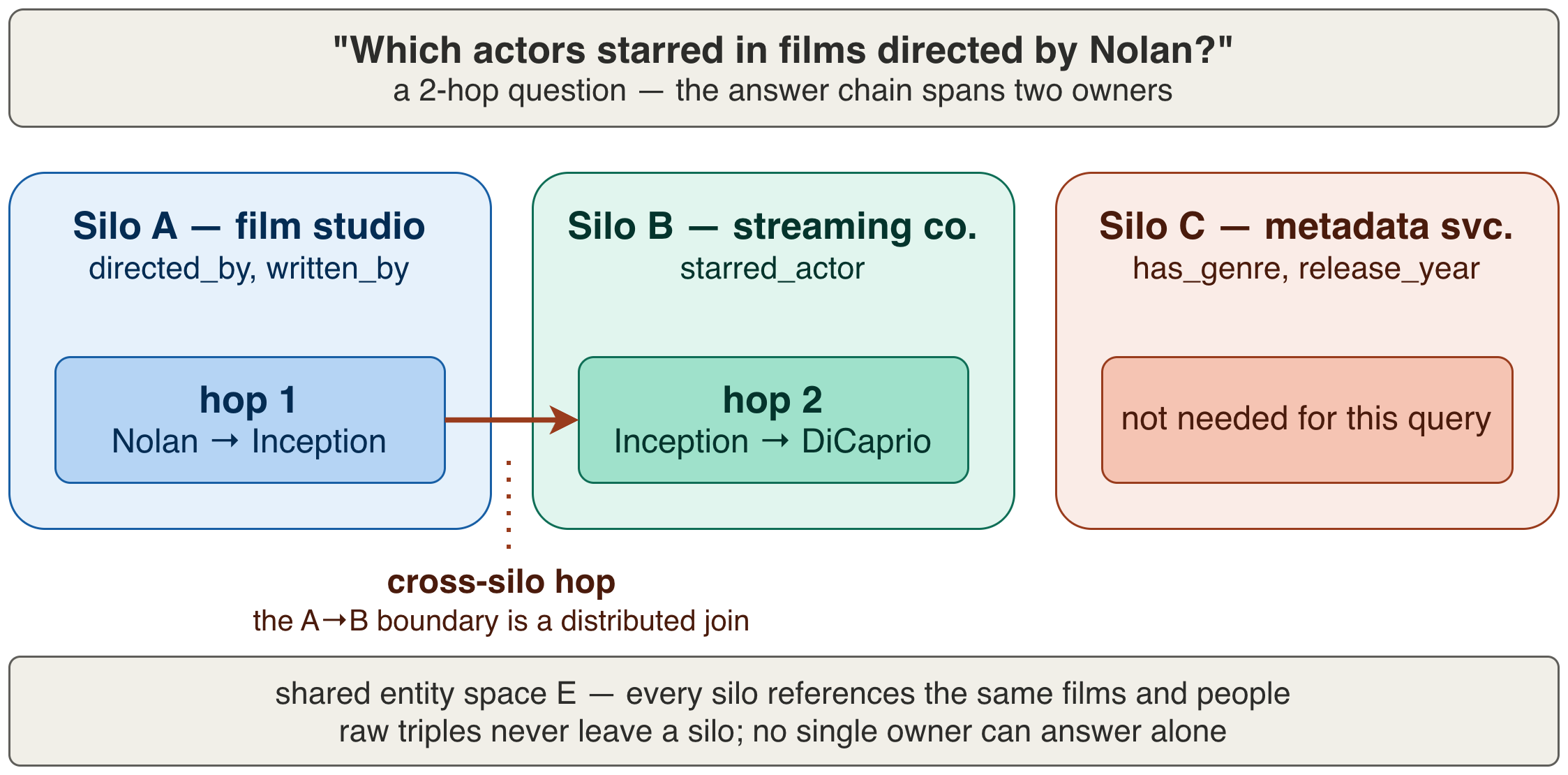}
\caption{Multi-hop question answering in the vertical federated setting.
Silos own disjoint relation subsets over a shared entity space $E$, so a
2-hop question may chain facts across a silo boundary.}
  \label{fig:motivating}
  \vspace{-12pt}
\end{figure}

Under vertical partitioning, that locality is lost. Answering a multi-hop
query means chaining facts that are split across parties by design: a
reasoning path may begin in one silo, pass through a shared entity, and end
in another. The question \emph{``Which actors starred in films directed by Nolan?''} requires the directing relation held by the studio and the acting relation held by the
platform, so neither party can answer it alone. Every hop that crosses a silo boundary plays the same role as a distributed join plays in a relational federation, since it forces evidence held by two owners to be combined (Figure~\ref{fig:motivating}). The
ingredients for answering such questions exist separately: embedding-based
methods rank answers to multi-hop questions over a centralized
graph~\cite{saxena2020improving}, and federated embedding methods learn
representations across parties without sharing raw
triples~\cite{chen2021fede}. Recent work has combined them to show that
multi-hop question answering is achievable over a vertically partitioned
graph, by training local embeddings within each silo and fusing them on a
server~\cite{bappy2026fedv}. That line of work, however, treats the partition as a fixed input
and focuses on answer quality.

\paragraph{A note on what each silo exposes:}
The vocabulary of this paper is drawn from data management, and terms such as distributed join and cross-silo hop describe the shape of the workload rather than the mechanism that serves it. No silo in our
setting exposes a query endpoint, and no silo evaluates a subquery over its own triples. A silo releases only derived quantities, here, a local entity embedding matrix. A question is answered by ranking candidate entities in a fused
embedding space, not by planning and executing a distributed query. This constraint is what makes the assignment of relations to silos a physical design decision with real consequences. There is no runtime optimizer that can
reorder, push down, or cache its way around a poor layout, so whatever a partition costs in cross-silo reasoning it costs on every query.

A prior and equally practical question has been largely overlooked: given a
relation vocabulary and a set of silos, \emph{how should the relations be
assigned to silos, and what storage and query costs does each assignment incur?}
This is squarely a data management problem. It concerns partitioning,
indexing, federated query processing, and the communication that federation
entails, and its answer is not cosmetic. Placing tightly co-used relations in the same silo shortens the cross-silo paths a query must traverse, but it can leave silos badly unbalanced in size, so that one overloaded silo throttles every synchronized round. Spreading relations evenly balances the load but separates relations that are frequently chained, lengthening cross-silo paths. A partition that is good on one axis is often poor on the other, at least
where each silo can hold several relations, and the same query workload can be cheap or
expensive to serve depending entirely on how the relations were divided.
These are exactly the trade-offs a practitioner must reason about when
deploying a federated knowledge graph, yet there is no systematic account
of them: a practitioner today chooses a partition, or inherits one, with no
way to know what it costs or how far it sits from the best achievable.

This paper provides that characterization. We treat partitioning as the
object of study rather than a fixed input, formalize it as an assignment of
the relation vocabulary to silos, and define a small space of strategies that
span the natural axes of the problem: semantic domain grouping, which reflects the
partition an organization typically inherits; frequency-balanced
partitioning, which equalizes the data held by each silo; co-occurrence
graph-cut partitioning, which keeps frequently chained relations together
to preserve locality; and random partitioning, which optimizes nothing and
serves as a baseline. We measure the consequences of each strategy along
five data management metrics, namely communication cost, candidate index
size, cross-silo path length, load balance, and end-to-end query latency.
Three of these prove to be invariants of the graph and the silo count,
leaving locality and balance as the two axes a partition actually trades
off. Crucially, we hold the learning model fixed, evaluating every strategy on
the same federated question answering substrate built on
TransE~\cite{bordes2013translating} and a frozen BERT
encoder~\cite{devlin2019bert}, so that at each silo count every difference in cost is attributable to the partition. Our contributions are as follows.

\begin{itemize}
  \item \textbf{Vertical partitioning as a design space.} We formalize the
  relation-to-silo assignment problem for vertically partitioned knowledge
  graphs and define four concrete strategies spanning the natural axes of
  the space.

\item \textbf{A data management cost model.} We define five measurable
  cost metrics, namely per-round communication, candidate set and index
  size, cross-silo path length, load balance, and query latency. Together
  they capture what a partition costs to store and query. We further show
  that three of the five are invariants of the graph and the silo count
  rather than consequences of the partition, so the partitioning decision
  reduces to a two-dimensional trade-off.

  \item \textbf{An empirical characterization.} Across two benchmarks and
  three silo counts, holding the learning model fixed, we quantify the trade-offs each strategy makes and distill practical guidance on which strategy
  suits which deployment objective.
\end{itemize}

\section{Related Work}
\label{sec:related}

\subsection{Partitioning and Physical Design for Graph Data}

How a graph dataset is laid out determines how expensive it is to query, and
choosing that layout has long been recognized as a core data management
decision. In RDF data management, SW-Store~\cite{abadi2009sw} introduced
vertically partitioned storage, grouping triples by predicate so that a query
touching only a few predicates scans little data. Balanced graph
partitioning remains an active area, and recent surveys catalogue the
maturity of the underlying
machinery~\cite{ccatalyurek2023more, ali2022survey}, while systems work
continues to improve partition quality at scale, as in streaming
partitioners that reduce both edge cuts and worker imbalance.
Most similar to our locality-oriented strategy is application-driven
partitioning~\cite{fan2023application}, which selects a layout from the
workload that will run over it rather than from graph topology alone.
Empirical comparisons of RDF partitioning
strategies~\cite{akhter2018empirical} confirm the recurring tension we also
study: partitioning to balance storage tends to sever the join paths that
queries traverse, so the layout minimizing imbalance is rarely the layout
minimizing cross-partition traffic, and no single scheme is uniformly best.

Our work adopts this physical design perspective but differs on two axes
that prior partitioning work does not combine. First, the query is a
\emph{natural language multi-hop question} answered by embedding-based
ranking, not a structured query with an explicit plan. The cost of a
partition is therefore mediated by a learned retrieval pipeline, namely
local embedding, server-side fusion, and candidate ranking, rather than by a
relational operator tree, and it is realized through structures such as the
per-silo candidate index that this pipeline builds. Second, the partition
boundary coincides with an \emph{ownership and privacy} boundary: relations
are held by autonomous organizations, raw triples cannot cross a silo, and a
cross-partition join is a round of federated communication between parties
that never expose their data rather than an intra-cluster shuffle. Together
these change both what a partition costs and which partitions are admissible. To
our knowledge, the cost of alternative relation-to-silo assignments has not
been characterized under these conditions, which is the gap this paper
addresses.

\subsection{Federated Knowledge Graphs and Question Answering}

Two further lines of work provide the substrate on which we measure these
costs, though neither studies the partitioning question itself. Federated
knowledge graph embedding learns representations across parties without
sharing raw data, but almost exclusively in the \emph{horizontal} setting,
where parties hold the same relations over different entities and a query
can often be answered within a single shard. FedE~\cite{chen2021fede}
established the pattern of aggregating entity embeddings through a server,
and the line has since developed rapidly, targeting heterogeneity and
unlearning~\cite{zhu2023heterogeneous, zhu2025parameter}, cheaper embedding
exchange~\cite{zhang2024low}, personalization~\cite{zhang2025personalized}, and the privacy and robustness risks of repeated exchange~\cite{hu2023quantifying, jiang2026unveiling}; benchmarks have also
matured~\cite{li2025openfgl}. This body of work shares our federated framing
but assumes the partition is given and targets model quality or privacy, not
the cost of the partition itself. The
vertical case, in which a single reasoning path is split across owners by
construction, remains comparatively unexplored. Vertical federation has been studied outside knowledge graphs, where parties hold disjoint feature sets over shared samples~\cite{tran2024differentially},
and federated learning is used more broadly where data cannot be centralized for
regulatory reasons~\cite{khan2024fed}. In both, the partitioned objects are
features rather than relations, so no reasoning path crosses a boundary.

Question answering over knowledge graphs has likewise advanced. Embedding-based methods score candidate answers in a learned space, as in
EmbedKGQA~\cite{saxena2020improving}, while recent work couples large
language models with graph traversal through agentic exploration, retrieved
relation paths, or multi-hop reasoning over evolving
graphs~\cite{sun2024think, luo2024reasoning, chen2024llm, ma2025large}. These
methods assume centralized graph access or centrally available retrieved
evidence. A smaller line extends
question answering to vertically partitioned federated graphs, establishing
that multi-hop answers can be recovered without centralizing the graph~\cite{bappy2026fedv}. That
work establishes feasibility and optimizes answer quality; it does not ask
how the underlying partition should be chosen or what different partitions
cost to store and query. We take such a pipeline as a \emph{fixed} substrate
and hold it constant, so that the differences we report are attributable to
the partition alone. Our focus is thus orthogonal and complementary: prior
work asks whether questions can be answered over a given partition, whereas
we ask how the partition should be chosen and what it costs.

\section{The Vertical Partitioning Design Space}
\label{sec:designspace}

\subsection{Problem Setup and Notation}

Let $G = (E, R, T)$ be a knowledge graph with entity set $E$, relation set
$R$, and triple set $T \subseteq E \times R \times E$. A \emph{vertical
partition} into $K$ silos is an assignment
$\Pi : R \rightarrow \{1, \dots, K\}$ that gives each silo $S_k$ a private
relation subset $R_k = \Pi^{-1}(k)$. The relation subsets are pairwise
disjoint and jointly cover $R$, so that
$R = \bigcup_{k=1}^{K} R_k$ and $R_i \cap R_j = \emptyset$ for $i \neq j$.
Silo $S_k$ holds only its local triples
$T_k = \{(h, r, t) \in T : r \in R_k\}$. The entity space $E$ is shared
across all silos, while the full triple set $T = \bigcup_k T_k$ is never
centralized. Raw triples cannot cross a silo boundary. As stated in Section~\ref{sec:intro}, a silo exposes only derived quantities, and in our substrate that means a local entity embedding matrix.

\begin{figure}[t]
  \centering
  \includegraphics[width=0.60\linewidth]{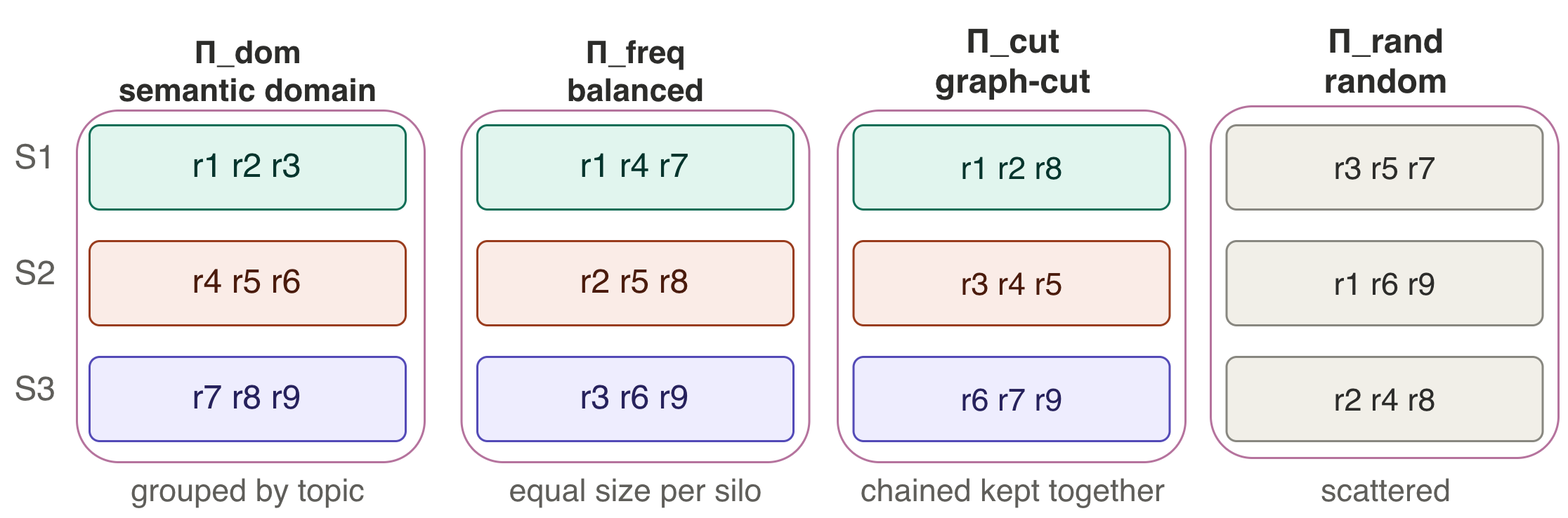}
\caption{The four partitioning strategies, shown for nine relations across
three silos: grouping by topic, equalizing triple counts, keeping chained
relations together, and random assignment.}
  \label{fig:strategies}
  \vspace{-9pt}
\end{figure}

The object of study is the assignment $\Pi$ itself. Two partitions of the
same graph into the same number of silos can differ substantially in silo
sizes, in the number of reasoning paths that cross a silo boundary, and in
the communication a federated query pipeline must perform, all before any
learning takes place. We consider four strategies that span the natural
axes of this space, illustrated in Figure~\ref{fig:strategies} for nine
relations across three silos.

\subsection{Semantic Domain Partitioning ($\Pi_{\mathrm{dom}}$)}

Relations are grouped by their semantic domain, mirroring how an
organization naturally owns a coherent slice of the graph, such as all
production relations in one silo and all cast relations in another. Each
relation $r$ carries a domain label $\mathrm{dom}(r)$, and
$\Pi_{\mathrm{dom}}$ treats each semantic domain as an indivisible group. When
domains outnumber silos they share silos; when silos outnumber domains, the
surplus silos remain empty. This models an inherited partition whose ownership
boundaries cannot be subdivided merely to occupy additional nodes. This is the assignment a real
federation typically inherits, because each participating organization
already owns a semantically coherent set of relations. It is intuitive and deployment-realistic, but it optimizes neither balance nor locality. Balance
suffers whenever domains differ in size, and severely so when silos outnumber
domains, because the surplus silos receives no relations. Locality suffers when two domains are frequently chained in queries, such as directing followed by acting, because
placing them in different silos forces every such query to cross a boundary.

\subsection{Frequency-Balanced Partitioning ($\Pi_{\mathrm{freq}}$)}

Relations are assigned so as to equalize the total triple count per silo. We
use a greedy longest-processing-time heuristic. Relations are sorted by their
triple count $|T_r|$ in descending order, and each is assigned in turn to
whichever silo is currently lightest. This drives the per-silo load toward
the ideal $|T| / K$, which directly optimizes load balance and storage
uniformity. What the heuristic does not consider is locality. Because it
weighs only size and never which relations are used together, it routinely
separates relations that queries chain, lengthening cross-silo reasoning
paths.

\subsection{Co-occurrence Graph-Cut Partitioning ($\Pi_{\mathrm{cut}}$)}

This strategy explicitly targets locality. We build a weighted relation
co-occurrence graph $G_R = (R, W)$ whose nodes are the relations. Its edge
weight $w(r_i, r_j)$ counts how often $r_i$ and $r_j$ appear, in either order, as consecutive
hops on a reasoning path in the training query workload:
\begin{equation}
  w(r_i, r_j) = \sum_{q} \mathbf{1}\big[\, r_i, r_j \text{ consecutive on the path of } q \,\big].
  \label{eq:cooc}
\end{equation}
The objective is to minimize the total weight of edges whose endpoints fall
in different silos,
\begin{equation}
  \min_{\Pi} \sum_{i < j} w(r_i, r_j) \cdot
  \mathbf{1}\big[\, \Pi(r_i) \neq \Pi(r_j) \,\big],
  \label{eq:cutobj}
\end{equation}
subject to a balance constraint that prevents one silo from absorbing most
relations.

We approximate Equation~\ref{eq:cutobj} with a greedy weighted-clustering
heuristic, which suits the vocabulary sizes in this setting, where
$|R| < 20$. Relations are processed in descending order of total incident
co-occurrence weight. Each is then placed in the silo maximizing the sum of
edge weights to relations already assigned there. Balance is controlled by a slack parameter $\epsilon$. A silo may hold at
most $\lfloor \epsilon \, |R| / K \rfloor$ relations, a cap relaxed only once
every silo has reached it. We set $\epsilon = 1.25$, permitting $25\%$ more
than the even share. Ties are broken toward the less populated silo, which
also places relations carrying no co-occurrence weight. Enrichment-derived
relations are treated as independent nodes. The partitioner may therefore
separate them from their base relations when the workload does not chain
them. Because $\Pi_{\mathrm{cut}}$ optimizes locality rather than size, it may
accept more imbalance than $\Pi_{\mathrm{freq}}$. Only $\Pi_{\mathrm{cut}}$
consults the query workload, and it does so through aggregate co-occurrence
counts over training paths.

\subsection{Random Partitioning ($\Pi_{\mathrm{rand}}$)}

Relations are assigned to silos uniformly at random. This strategy optimizes
nothing and serves as the reference point against which the other three are
measured, since a strategy that cannot beat $\Pi_{\mathrm{rand}}$ on the
metric it targets is not earning its complexity. A single random draw may
happen to be favorable or unfavorable. We therefore run $\Pi_{\mathrm{rand}}$
with twenty seeds for the structural metrics and three for the metrics that
require training, and report the mean across them. Together the four strategies span the design space, from the inherited
default through the balance-optimal and locality-optimal extremes to an
assignment that optimizes neither.

\begin{figure}[t]
  \centering
  \includegraphics[width=0.60\linewidth]{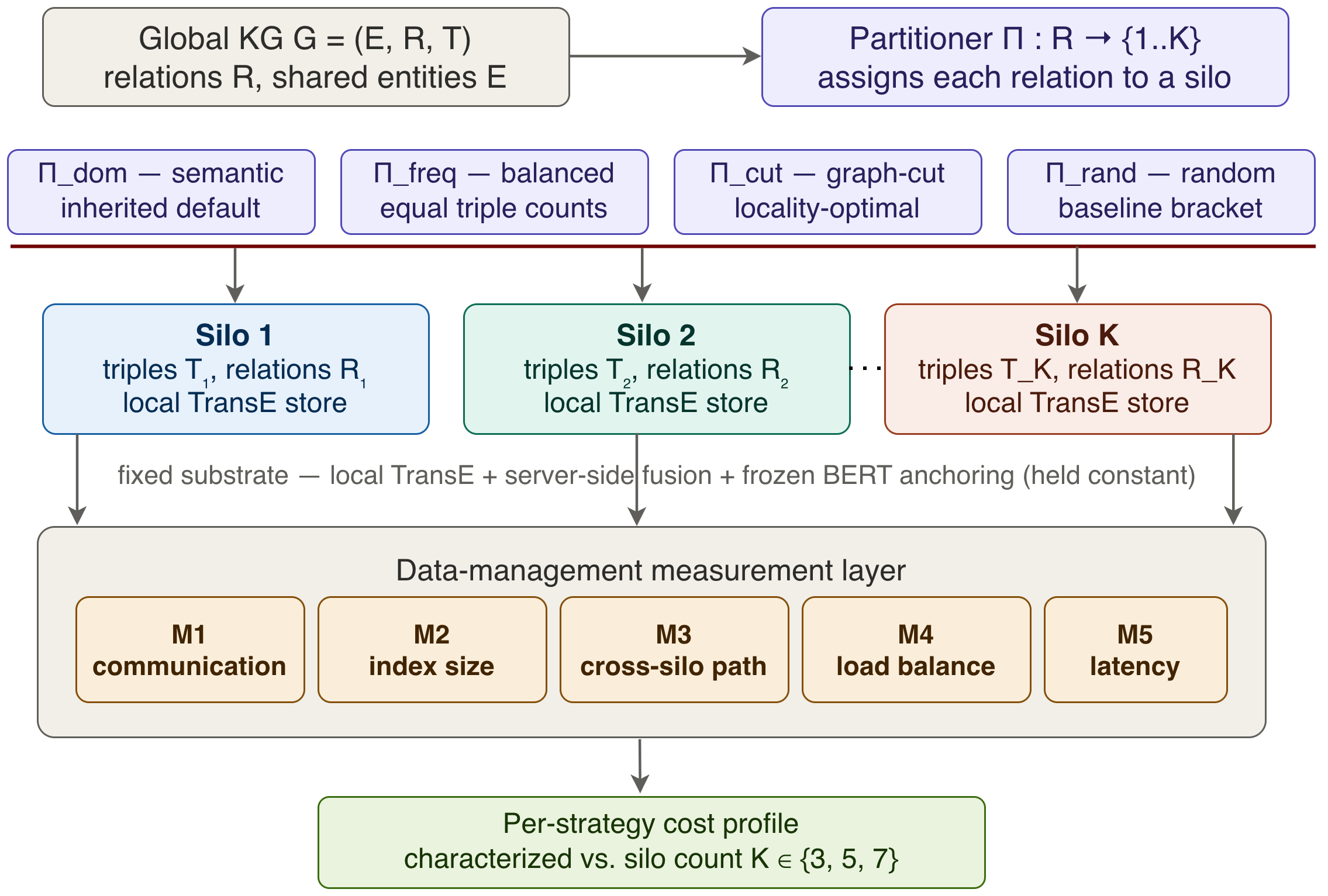}
  \caption{Vertical partitioning as a data-management decision. A strategy
$\Pi$ assigns each relation to one of $K$ silos. The resulting layout is
trained on a fixed substrate and characterized along five cost metrics
(M1--M5) for $K \in \{3,5,7\}$, so that, within each $K$, differences reflect the partition.}
  \label{fig:methodology}
  \vspace{-9pt}
\end{figure}

\subsection{Computing the Partitions}

All four assignments are computed offline, once, before any training.
$\Pi_{\mathrm{rand}}$ and $\Pi_{\mathrm{dom}}$ are $O(|R|)$,
$\Pi_{\mathrm{freq}}$ is $O(|R| \log |R|)$, and $\Pi_{\mathrm{cut}}$ requires
one pass over the training paths followed by a balanced $K$-way cut over $|R|$
nodes. Since relation vocabularies are small, partitioning is a one-time cost
negligible against the recurring costs of Section~\ref{sec:metrics}, and we
exclude it from the comparisons.

\section{Cost Metrics and Methodology}
\label{sec:metrics}

\begin{figure}[t]
  \centering
  \includegraphics[width=0.62\linewidth]{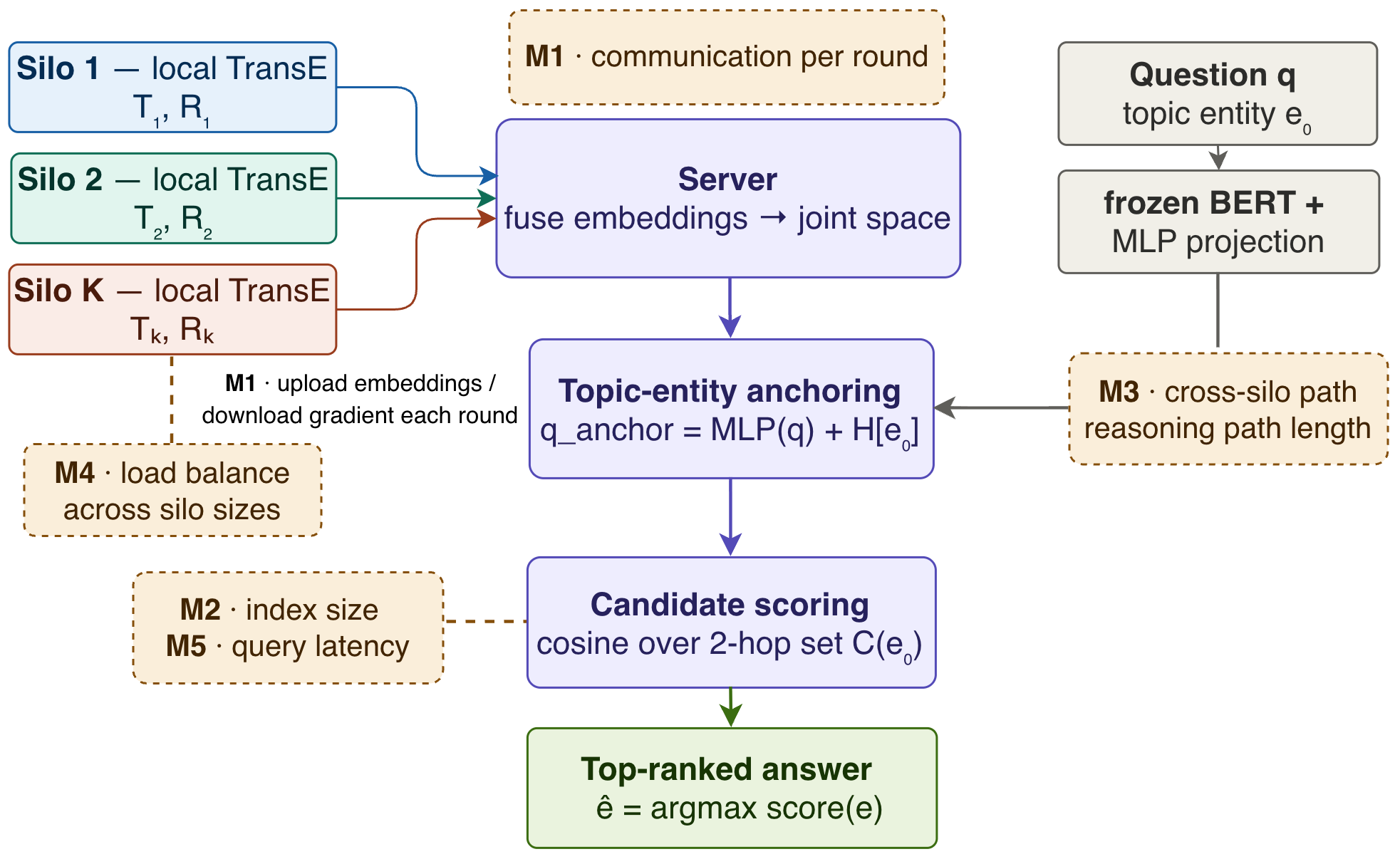}
 \caption{The fixed federated question-answering substrate used in all
experiments. Each silo trains a local TransE model and uploads entity
embeddings. The server fuses them, anchors the question at the topic entity,
and scores candidates over a two-hop set. Dashed annotations mark where each
cost metric (M1--M5) is incurred.}
  \label{fig:substrate}
  \vspace{-9pt}
\end{figure}

We characterize each partition along five data-management metrics, measured
on a fixed federated question answering substrate
(Figure~\ref{fig:substrate}). Holding the substrate constant is what lets
us attribute every difference to the partition alone. Lower is better for all five metrics. Table~\ref{tab:metrics} summarizes the five metrics, the units in which each
is reported, and whether it varies with the partition.

\begin{table}[t]
\caption{The five cost metrics, the units in which each is reported, and
whether it varies with the partition. M3 and M4 are the two axes a partition
trades off. The other three are fixed by the graph, the substrate, and the
silo count.}
\label{tab:metrics}
\centering
\small
\begin{tabular}{llllc}
\toprule
ID & Name & Reported as & Definition & Depends on $\Pi$? \\
\midrule
M1 & Communication cost & GB per round & Eq.~\ref{eq:comm} & No \\
M2 & Candidate set and index size & Entries per topic entity & Eq.~\ref{eq:index} & No \\
M3 & Cross-silo path length & Crossing rate per query & Eq.~\ref{eq:cspl} & Yes \\
M4 & Load balance & Imbalance, CV of silo sizes & Eq.~\ref{eq:imbalance} & Yes \\
M5 & Query latency & Mean ms per question & --- & No \\
\bottomrule
\end{tabular}
\end{table}

\textbf{The fixed substrate.} Every strategy is evaluated on the same
pipeline, summarized in Figure~\ref{fig:substrate} and held constant across
all experiments. Each silo $S_k$ trains a local knowledge graph embedding on
its own triples $T_k$ using TransE~\cite{bordes2013translating}, and uploads
only its entity embedding matrix to a server; raw triples and relation
embeddings never leave the silo. The server fuses the per-silo entity
embeddings into a joint representation. A natural language question is encoded by a frozen BERT
encoder~\cite{devlin2019bert} followed by a trainable projection, anchored at
the topic entity, and used to score candidate answers by similarity in the
joint space. At each $K$, the partition is therefore the only independent variable, and
the co-occurrence statistics needed by $\Pi_{\mathrm{cut}}$ are the only
workload information any component reads.

\textbf{M1: Communication cost.} Federated training proceeds in rounds. Each
round, every silo uploads its entity embedding matrix and receives a gradient
slice in return, so the per-round volume is
\begin{equation}
  C_{\mathrm{round}} = K \cdot |E| \cdot d \cdot 2 \cdot 4 \text{ bytes},
  \qquad
  C_{\mathrm{total}} = T \cdot C_{\mathrm{round}},
  \label{eq:comm}
\end{equation}
where $|E|$ is the shared entity count, $d$ the embedding dimension, the
factor $2$ covers upload and returned gradient, and $4$ the bytes per float.
$C_{\mathrm{round}}$ is therefore identical for every partition at a given
$K$, and we report it as M1. $C_{\mathrm{total}}$ also depends on $T$, the rounds needed to reach the
validation-quality target. M1 measures training-time communication only;
inference contacts no silo in the evaluated substrate; M3 is therefore reported as a structural locality measure rather than as measured inference communication. It estimates the fragmentation that a partition would impose on systems that execute relation-local reasoning across owners.

\textbf{M2: Candidate set and index size.} To answer a question without
traversing the graph at query time, the substrate precomputes, for each topic
entity $e_0$, a set of candidate answer entities reachable within two hops.
A two-hop chain may have its first hop in one silo and its second in another.
The candidate set is therefore computed over the pooled neighborhood across
silos, since expanding each silo independently and taking the union would
miss exactly the cross-silo chains this setting is defined by. We report its
mean size,
\begin{equation}
  \mathrm{IndexSize} = \frac{1}{|E_{\mathrm{topic}}|}
  \sum_{e_0 \in E_{\mathrm{topic}}} \big| C(e_0) \big|,
  \label{eq:index}
\end{equation}
where $C(e_0)$ is the pooled two-hop candidate set of $e_0$. This set depends on the graph rather than on $\Pi$, making it
a control that confirms every strategy ranks over an equivalent candidate
pool. 

\textbf{M3: Cross-silo path length.} Cross-silo path length (CSPL) is the direct
measure of locality. For each query, its reasoning path is a sequence of
relations $r_1, r_2, \dots$, and a hop crosses a silo boundary whenever two
consecutive relations are assigned to different silos. We report the mean
number of such crossings per query,
\begin{equation}
  \mathrm{CSPL} = \frac{1}{|Q|} \sum_{q \in Q}
  \sum_{i} \mathbf{1}\big[\, \Pi(r_i) \neq \Pi(r_{i+1}) \,\big].
  \label{eq:cspl}
\end{equation}
A path spanning two relations contains one transition, contributing $0$ when
both relations lie in the same silo and $1$ otherwise. CSPL counts crossings
per query and is bounded by 1 on two-hop workloads, so its value equals the
fraction of queries that cross a boundary, and we report it as a crossing rate throughout. Enrichment creates direct relations between entities originally two hops apart, so recovery returns a single-relation path for 37.9\% of MetaQA and 36.7\% of PathQuestion questions; these contain no transition and contribute
zero. CSPL is computed over all recovered paths. A cross-silo hop is the
structural analogue of a distributed join. Minimizing CSPL is the objective of
$\Pi_{\mathrm{cut}}$ (Equation~\ref{eq:cutobj}), which builds its co-occurrence
graph from the same relation pairs this metric counts.

\textbf{M4: Load balance.} Federation proceeds in synchronized rounds, so the
heaviest silo bounds each round and an unbalanced partition wastes the
capacity of the lighter ones. We quantify imbalance as the coefficient of
variation (CV) of the per-silo triple counts,

\begin{equation}
  \mathrm{Imbalance} = \frac{\sigma\big(\{|T_k|\}_{k=1}^{K}\big)}
  {\mu\big(\{|T_k|\}_{k=1}^{K}\big)},
  \label{eq:imbalance}
\end{equation}
where $\mu$ and $\sigma$ are the mean and standard deviation of the silo
sizes. M4 is reported as imbalance rather than balance so that lower is better for all five metrics. This is the metric $\Pi_{\mathrm{freq}}$ is designed to minimize. Where the relation budget per silo is generous, it stands in opposition to CSPL.

\textbf{M5: Query latency.} Query latency is the end-to-end time to answer a
single question, comprising candidate lookup, question encoding, topic
anchoring, and scoring. We report mean query latency. Latency is downstream of M2, so it should not vary with the partition beyond measurement noise.

\textbf{Two families of cost.} Only M3 and M4 are consequences of the
partition. M2 and M5 are determined by the graph and the substrate. Pooled
candidate construction makes the scored set independent of $\Pi$ by design, and latency is downstream of it. M1, the per-round volume of
Equation~\ref{eq:comm}, is fixed by the silo count. Total bytes also depend on
the round count $T$, which varies across training runs
(Section~\ref{sec:results}). We therefore verify M1, M2, and M5 as invariants
and treat partitioning as a two-axis problem in locality and balance. How far
those axes pull apart is what the study measures.

\section{Experimental Setup}
\label{sec:setup}

\textbf{Datasets.} We evaluate on two multi-hop knowledge graph question
answering benchmarks that differ in domain, in graph density, and in the
size of the relation vocabulary being partitioned.
MetaQA~\cite{zhang2018variational} is a movie domain benchmark built over a
WikiMovies knowledge base, with questions at one, two, and three hops; we
use the two-hop split. PathQuestion~\cite{zhou2018interpretable} is a person-centric benchmark derived from Freebase, covering family, demographic, and
biographical relations; we use its two-hop split. Both graphs are enriched
offline with inverse and property chain axioms before partitioning, so that
answer entities are reachable from the topic entity within the two-hop
expansion the substrate performs. This enrichment is applied identically in
every experimental cell and therefore does not confound the comparison. It
also expands the relation vocabulary that is subsequently partitioned:
MetaQA grows from nine original relations to fifteen, while PathQuestion remains at
thirteen relations, since its enrichment materializes inverse triples under existing
relation names.

Table~\ref{tab:datasets} summarizes the statistics that interact with
partitioning. The relation count determines the size of the design space,
while the entity count drives per-round communication and index size. The
two benchmarks stress different parts of the cost model. PathQuestion has a
larger entity space but fewer triples, so it is the sparser graph, whereas
MetaQA is denser over fewer entities.

\begin{table}[t]
  \centering
  \small
   \renewcommand{\arraystretch}{0.80}
 \caption{Benchmark statistics relevant to partitioning. The relation count is
the size of the set being partitioned. The entity count drives communication
volume and index size.}
  \label{tab:datasets}
  \begin{tabular}{@{}lrrrl@{}}
    \toprule
    Dataset & Entities & Relations & Triples & Domain \\
    \midrule
    MetaQA        & 43{,}235 & 15 & 405{,}433 & movie \\
    PathQuestion  & 75{,}043 & 13 & 376{,}847 & person-centric \\
    \bottomrule
  \end{tabular}
\end{table}

\textbf{Fixed learning substrate.} Because this is a study of partitioning,
the learning model is held constant. We fix TransE~\cite{bordes2013translating}
as the knowledge graph embedding and a frozen BERT~\cite{devlin2019bert}
question encoder with a small trainable projection head. TransE keeps candidate
scoring inexpensive and is a stable, widely used baseline. Fixing this pair
isolates the effect of the partition, so any difference in the five metrics is
attributable to $\Pi$. We therefore do not sweep alternative embeddings or
encoders.

\textbf{Training configuration.} Every cell uses the same training configuration. Local
TransE models use 256-dimensional embeddings trained with a margin ranking
loss, margin $1.0$, ten negative samples per triple, Adam with learning rate
$10^{-3}$, and batch size $512$. On the server, the BERT encoder is frozen
and only a two-layer projection head is trained, using Adam with learning
rate $10^{-4}$, batch size $64$, and margin $1.0$. Gradient norms are
clipped to $1.0$, and entity embeddings are renormalized to the unit sphere 
after each update. Candidate construction expands two hops from each topic
entity, with neighbor caps of $50$ at the first hop and $20$ at the second,
and at most $100$ neighbors retained per entity. The only quantities that
change across cells are the partition $\Pi$ and the silo count $K$.

\textbf{Partitions and silo counts.} We apply each of the four strategies
$\Pi_{\mathrm{dom}}$, $\Pi_{\mathrm{freq}}$, $\Pi_{\mathrm{cut}}$, and
$\Pi_{\mathrm{rand}}$ at $K \in \{3, 5, 7\}$ silos. Each relation is assigned
to exactly one silo, and the entity space is shared across all silos.
$\Pi_{\mathrm{rand}}$ is averaged over twenty seeds for the structural
metrics M3 and M4, which require no training, and over three seeds for the
quality control and total communication to target, which each require a full training run. This yields a
matrix of four strategies by three silo counts for each dataset.

\textbf{Recovering reasoning paths.} Both benchmarks provide a question, a
topic entity, and an answer entity. Neither annotates the relations traversed
between them. Since M3 and the co-occurrence graph of $\Pi_{\mathrm{cut}}$
are both defined over relation sequences, we recover a reasoning path for
each question by breadth-first search over the global graph, taking the
shortest path from topic entity to answer. Reverse traversals are normalized
to their base relation, so the co-occurrence graph, the partition, and the
cross-silo path length metric all use the same relation identities. Recovery
runs separately on the training and testing splits. $\Pi_{\mathrm{cut}}$ consults only
training paths, through aggregate co-occurrence counts, while M3 is measured
over held-out test paths the partitioner never sees. Recovery yields
aggregate counts, so no individual query needs to leave a silo. When several
distinct paths connect a topic entity to an answer, the shortest is a proxy
for the intended reasoning chain, which is a limitation of the analysis.

\textbf{Protocol.} For each combination of dataset, strategy, and silo count,
we build the partition, construct the candidate index offline, train the
fixed substrate, and record M1 through M5. The quality target is Hits@3 on the validation split,
set per dataset to reflect the achievable ceiling: $0.70$ on MetaQA and
$0.55$ on PathQuestion, whose smaller training set supports a lower one.
Training runs for 40 rounds on MetaQA and 100 on PathQuestion, and the round first crossing the target defines $T$ in Equation~\ref{eq:comm}. Most
cells clear the target early, so differences in $T$ are small and may reflect run-to-run variation. Quality serves as a control rather than a result we claim. All measurements
use a single NVIDIA H100 GPU.

\textbf{Reproducibility.} The partitioning algorithms, the candidate index
construction, and the measurement harness are independent of the learning
substrate, so the characterization can be
reproduced\footnote{Code: \url{https://github.com/brains-group/vertical-fkg-partitioning}}
with any embedding or encoder. 
Each cell is determined by the dataset, the
strategy, the silo count, and, for $\Pi_{\mathrm{rand}}$, the random seed.

\section{Results}
\label{sec:results}

Table~\ref{tab:results} reports the two metrics that differ across
strategies, together with the quality control. We first confirm comparable
answer quality, then examine locality and balance.

\begin{table}[t]
\caption{Partitioning cost results for the two metrics that depend on the
partition. M3 is the cross-silo path length of Equation~\ref{eq:cspl},
reported as a crossing rate and measured on held-out test paths. M4 is the
load balance metric of Equation~\ref{eq:imbalance}, reported as the
coefficient of variation of per-silo triple counts. Lower is better for both.
Hits@3 on the test split is a quality control, not a result we claim.
$\Pi_{\mathrm{rand}}$ is averaged over twenty seeds for M3 and M4 and over
three seeds for quality. The best M3 and M4 value in each row is in bold.}
\label{tab:results}
\centering
\small
\begin{tabular}{llcccccccc}
\toprule
& & \multicolumn{4}{c}{MetaQA} & \multicolumn{4}{c}{PathQuestion} \\
\cmidrule(lr){3-6} \cmidrule(lr){7-10}
Metric & $K$
& $\Pi_{\mathrm{dom}}$ & $\Pi_{\mathrm{freq}}$ & $\Pi_{\mathrm{cut}}$ & $\Pi_{\mathrm{rand}}$
& $\Pi_{\mathrm{dom}}$ & $\Pi_{\mathrm{freq}}$ & $\Pi_{\mathrm{cut}}$ & $\Pi_{\mathrm{rand}}$ \\
\midrule
\multirow{3}{*}{\shortstack[l]{M3: Cross-silo path\\length (crossing rate)}}
 & 3 & 0.019 & 0.058 & \textbf{0.000} & 0.048 & 0.517 & 0.379 & \textbf{0.276} & 0.394 \\
 & 5 & 0.019 & 0.019 & \textbf{0.000} & 0.053 & 0.517 & \textbf{0.448} & 0.517 & 0.476 \\
 & 7 & \textbf{0.019} & \textbf{0.019} & \textbf{0.019} & 0.060 & 0.517 & 0.535 & 0.517 & \textbf{0.503} \\
\midrule
\multirow{3}{*}{\shortstack[l]{M4: Load balance\\(imbalance, CV)}}
 & 3 & 0.371 & \textbf{0.018} & 0.326 & 0.428 & 0.609 & \textbf{0.012} & 0.168 & 0.321 \\
 & 5 & 0.946 & \textbf{0.302} & 0.619 & 0.656 & 1.134 & \textbf{0.021} & 0.661 & 0.513 \\
 & 7 & 1.286 & \textbf{0.519} & 0.537 & 0.745 & 1.483 & \textbf{0.206} & 0.409 & 0.560 \\
\midrule
\multirow{3}{*}{\shortstack[l]{Quality control:\\Hits@3}}
 & 3 & 0.814 & 0.856 & 0.828 & 0.841 & 0.800 & 0.728 & 0.831 & 0.764 \\
 & 5 & 0.799 & 0.817 & 0.816 & 0.811 & 0.800 & 0.754 & 0.728 & 0.740 \\
 & 7 & 0.796 & 0.810 & 0.801 & 0.808 & 0.661 & 0.774 & 0.749 & 0.727 \\
\bottomrule
\end{tabular}
\end{table}

\textbf{Quality is matched on MetaQA, loosely on PathQuestion.} Cost comparisons are meaningful only at
matched quality. On MetaQA, Hits@3 spans $0.796$ to $0.856$. The spread at a
fixed $K$ is at most 
0.042. On PathQuestion it spans $0.661$ to $0.831$.
The wider range reflects a much smaller training set of $1{,}524$ questions.
On PathQuestion, the other strategies change rank as $K$ changes, while $\Pi_{\mathrm{rand}}$, averaged over three seeds, stays third at every $K$. This is what we would expect if the wider spread reflects run-to-run variation on a small training set rather than the partition. We therefore treat PathQuestion as loosely matched rather than matched, so cost comparisons on that benchmark
should be read as approximate. Quality is reported only to confirm that the strategies are comparable, not as a result we claim.

\textbf{Locality.} Crossing rates are measured on held-out test paths, while
$\Pi_{\mathrm{cut}}$ builds its co-occurrence graph from training paths only.
The numbers therefore reflect generalization to unseen queries rather than
fit to the optimization target. On MetaQA, $\Pi_{\mathrm{cut}}$ records
$0.000$ at $K=3$ and $K=5$, against $0.019$ for $\Pi_{\mathrm{dom}}$ and
$0.048$ and $0.053$ for $\Pi_{\mathrm{rand}}$, respectively, and ties for best at $K=7$. Absolute
magnitudes are small there. Every strategy stays at or below $0.06$, so
partitioning has limited practical effect on locality for this benchmark.

PathQuestion shows a much larger locality effect. At $K=3$, $\Pi_{\mathrm{cut}}$
records $0.276$ against $0.517$ for the inherited partition, a reduction of
$47\%$. The contrast sharpens on paths that contain a relation transition,
where a crossing is possible at all. Shortest-path recovery returns a
single-relation path for $36.7\%$ of PathQuestion questions, leaving $63.3\%$
where transitions are possible. Dividing by that fraction, the inherited
partition crosses a boundary on $82\%$ of such queries and
$\Pi_{\mathrm{cut}}$ on $44\%$. At $K=5$ and $K=7$, $\Pi_{\mathrm{cut}}$ no longer leads. As $K$ grows, each silo holds fewer
of the $13$ to $15$ relations, and the balance cap admits at most
$\lfloor 1.25\,|R|/K \rfloor$ relations per silo. At $K=7$ this leaves room
for two, too little freedom to co-locate a chain. $\Pi_{\mathrm{dom}}$
records the same crossing rate at every $K$ on both benchmarks, because the
partition itself does not change. With three semantic domains, silos beyond
the third receive no relations, and the relation-to-silo map is identical at
$K=3$, $5$, and $7$.

\begin{figure}[t]
  \centering
  \includegraphics[width=0.65\linewidth]{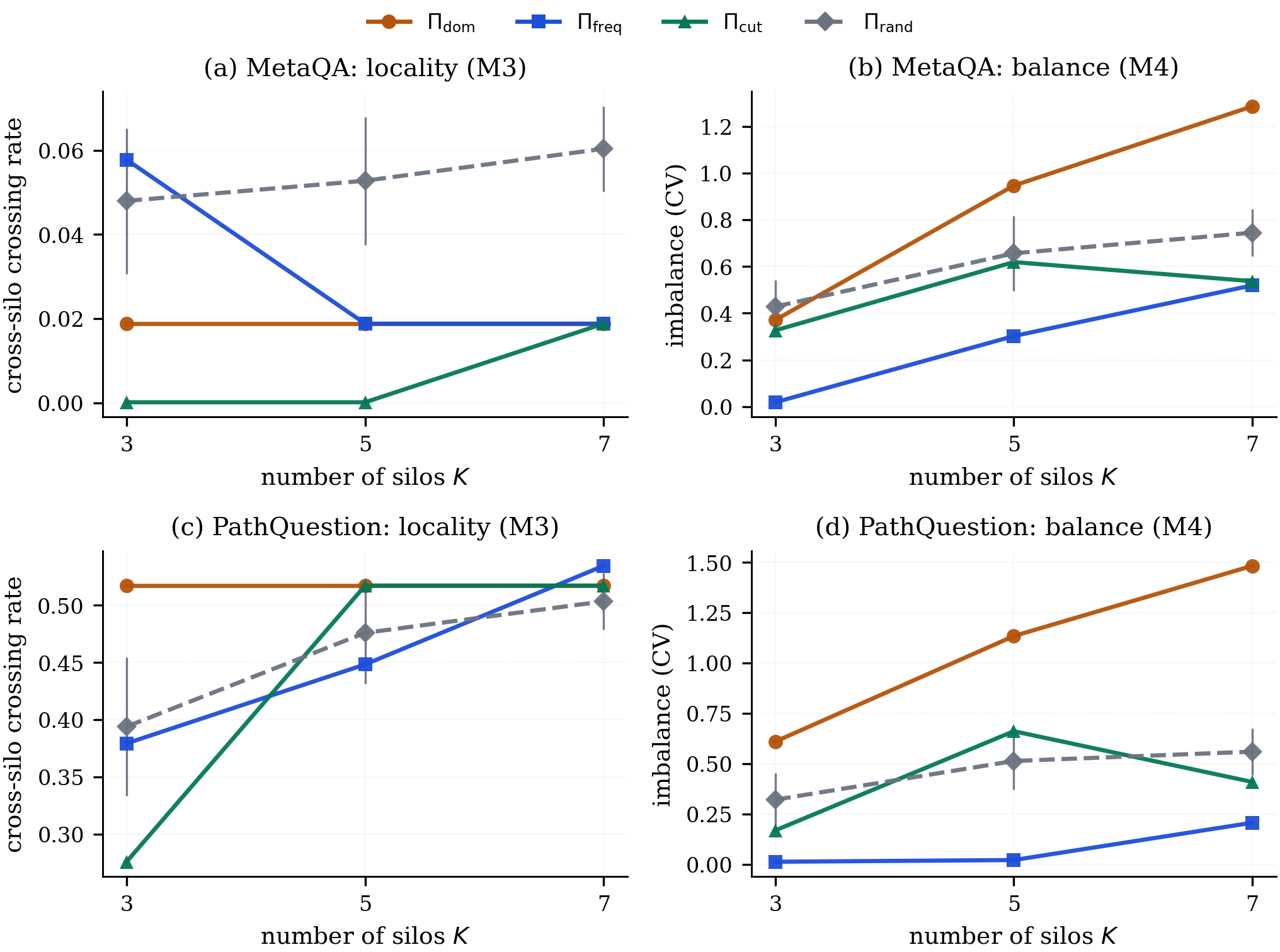}
  \caption{Cost metrics versus silo count $K$. Panels (a,c) show crossing rate
(M3) and panels (b,d) load imbalance (M4); lower is better. Crossing rates use
held-out test paths, and $\Pi_{\mathrm{rand}}$ is averaged over twenty seeds.
$\Pi_{\mathrm{cut}}$ leads on locality at $K=3$, while
$\Pi_{\mathrm{freq}}$ consistently minimizes imbalance. Panels (a) and (c)
use different vertical scales.}
  \label{fig:cost_curves}
  \vspace{-9pt}
\end{figure}

\textbf{Balance.} $\Pi_{\mathrm{freq}}$ achieves the lowest imbalance in all
six combinations of dataset and silo count. At $K = 3$ it records 0.018 on MetaQA and 0.012 on PathQuestion, roughly twenty-four and twenty-seven times lower than $\Pi_{\mathrm{rand}}$. $\Pi_{\mathrm{dom}}$ becomes the worst-balanced strategy at $K \geq 5$ on both
datasets and deteriorates sharply as $K$ grows, from $0.371$ to $1.286$ on
MetaQA and from $0.609$ to $1.483$ on PathQuestion. The cause is again the empty surplus silos, which inflate the coefficient of variation, and it is a concrete limitation for practitioners
scaling out an existing federation.
Figure~\ref{fig:cost_curves} shows both trends across $K$.

\begin{figure}[t]
  \centering
  \includegraphics[width=0.65\linewidth]{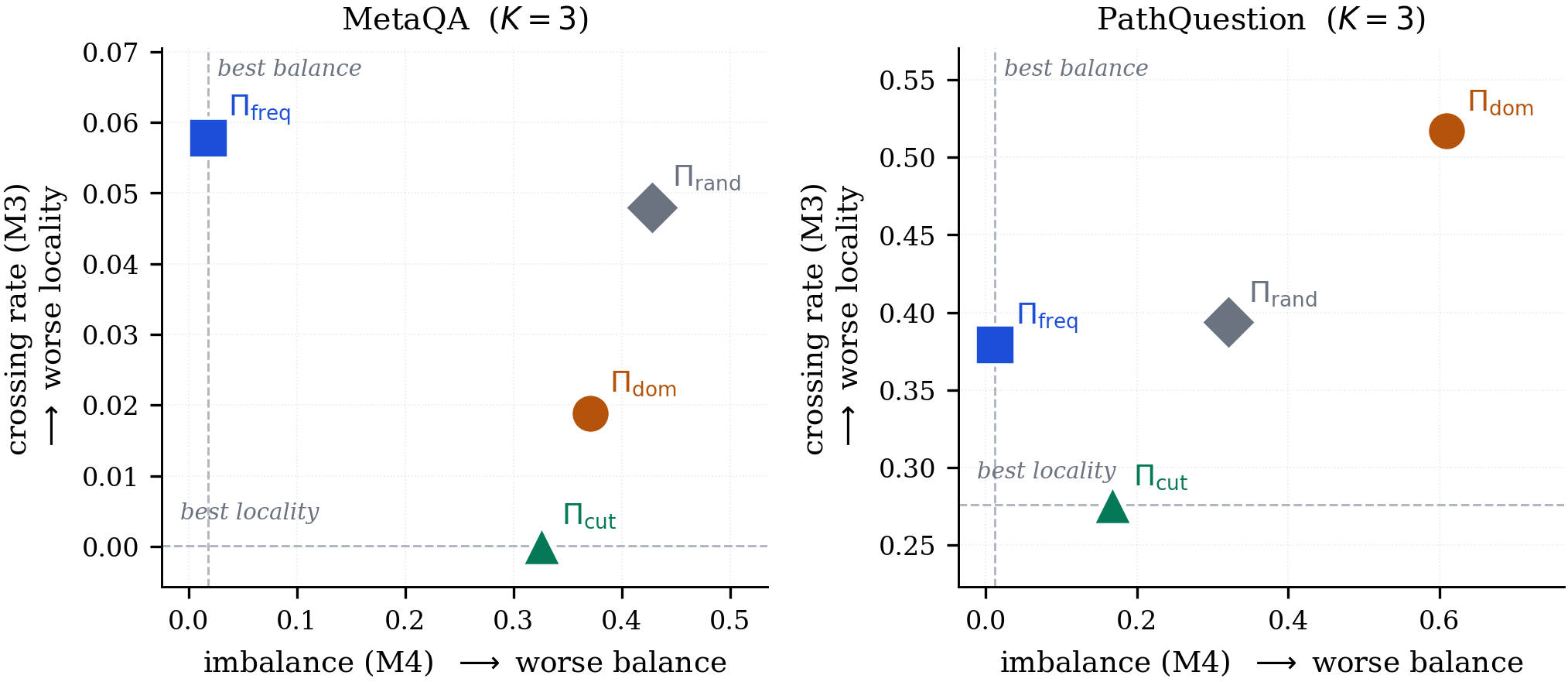}
  \caption{Locality against balance at $K=3$, where the trade-off is
  sharpest. Dashed lines mark the best value achieved on each axis, and no
  strategy reaches their intersection. $\Pi_{\mathrm{cut}}$ is lowest on
  crossing rate and $\Pi_{\mathrm{freq}}$ is leftmost on imbalance, and
  no strategy is best on both axes. At $K \geq 5$ the opposition weakens
  (Figure~\ref{fig:cost_curves}).}
  \label{fig:tradeoff}
  \vspace{-9pt}
\end{figure}

\textbf{Two regimes rather than a universal trade-off.} The relationship
between locality and balance depends on the ratio of relations to silos. At
$K=3$, where each silo can hold several relations, the two objectives
conflict directly. On MetaQA, $\Pi_{\mathrm{cut}}$ attains the best crossing
rate in the study, $0.000$, but does so at $0.326$ imbalance.
$\Pi_{\mathrm{freq}}$ attains the best imbalance, $0.018$, at the highest
crossing rate of any non-random strategy, $0.058$. Neither dominates the
other, and Figure~\ref{fig:tradeoff} shows no strategy in the lower-left
corner.

At larger values of $K$ the pattern changes. As the number of silos increases, each holds fewer of the
$13$ to $15$ relations and the balance cap admits fewer co-located pairs.
Locality differences collapse while imbalance differences persist. On PathQuestion, $\Pi_{\mathrm{freq}}$ is at least as good as $\Pi_{\mathrm{cut}}$ on \emph{both} axes at $K=5$; at $K=7$ it gives up $0.018$ on crossing rate while holding roughly half the imbalance. The trade-off therefore weakens as $K$ grows. Where the relation budget per silo is thin, balance-aware partitioning is the safer choice because it retains a strong load-balance advantage while remaining competitive on locality.

\textbf{Remaining metrics.} The other three metrics behave as the cost model
predicts. Candidate index size (M2) is constant across strategies, at $144.3$
entries per topic entity on MetaQA and $103.6$ on PathQuestion. Pooled
candidate construction makes the scored set depend on the graph rather than
on $\Pi$, which confirms that all strategies rank over an equivalent pool.
Query latency (M5) is correspondingly comparable, averaging $10.5$~ms on
MetaQA and $10.0$~ms on PathQuestion. Per-round communication (M1) follows
Equation~\ref{eq:comm} and is fixed by the silo count, at $0.27$, $0.44$, and
$0.62$~GB per round on MetaQA for $K=3$, $5$, and $7$. Total bytes to target
depend on the round count $T$, which varies across cells without a systematic
pattern. Across three training seeds of $\Pi_{\mathrm{rand}}$ at $K = 5$, total bytes to target vary by a factor of two, a range as wide as that between strategies.
Differences in convergence are therefore within training variance, not a clear
effect of the partition. A partition should be chosen for locality or balance, not to
reduce training bandwidth or query time.

\textbf{What is structural and what is measured.} Two of our observations follow from the setup rather than from data. That $\Pi_{\mathrm{dom}}$ leaves silos empty once $K$ exceeds the number of semantic domains is a property of the strategy, and that the balance cap admits at most $\lfloor \epsilon |R| / K \rfloor$ relations per silo is arithmetic. We report them because a practitioner scaling out an inherited federation will encounter them, not as findings. What is measured is where the two axes fall for a given graph and workload. The 47\% reduction in crossing rate on PathQuestion at $K = 3$ is
empirical, as is the fact that $\Pi_{\mathrm{freq}}$ matches
$\Pi_{\mathrm{cut}}$ on locality at $K = 5$ 
with roughly 31-fold 
lower imbalance. Neither is predictable from the framework. The quantity that decides which regime holds is relations per silo, $|R| / K$, not $|R|$ or $K$
alone. Our benchmarks reach the thin regime at $K = 7$ only because $|R|$ is small, and a federation with two hundred relations would reach it near $K = 100$. We expect the guidance to transfer by that ratio, but we have not
verified it on a large vocabulary and mark this as the main open item.

\section{Conclusion}
\label{sec:conclusion}

We treated the vertical partition as an object of study rather than a fixed
input, formalizing the relation-to-silo assignment as a design space and
defining five cost metrics. Three proved to be invariants of the graph and
the silo count, leaving locality and balance as the axes a partition
trades off. Locality-aware partitioning works where the relation budget
allows it: at $K=3$ on PathQuestion, $\Pi_{\mathrm{cut}}$ cuts the crossing
rate from $0.517$ to $0.276$. Balance-aware partitioning works reliably,
with $\Pi_{\mathrm{freq}}$ lowest on imbalance in every cell. The conflict is
strongest at low silo counts and weakens as $K$ grows; once each silo holds
too few relations to co-locate a chain, $\Pi_{\mathrm{freq}}$ becomes
competitive on locality while keeping its advantage on balance. Guidance is
conditional. Federations with room to co-locate reasoning paths should favor
$\Pi_{\mathrm{cut}}$; those bounded by their slowest silo, or spread across
many silos, should favor $\Pi_{\mathrm{freq}}$. An inherited partition sits
far from both optima and does not populate more silos than there are semantic domains. 
Three limitations bound these claims. Our benchmarks have fewer than twenty relation types, so the thin regime arrives at $K = 7$. Since the governing
quantity is relations per silo, we expect the same two regimes at proportionally larger silo counts on a large vocabulary, but this remains a conjecture. $\Pi_{\mathrm{cut}}$ also uses a greedy heuristic over recovered
rather than annotated paths, which biases M3 and the method it evaluates in the same direction. Finally, we characterize four heuristics rather than optimizing the trade-off directly; a partitioner minimizing a weighted combination of cut weight and imbalance would trace the frontier between the
two axes. Larger vocabularies, a frontier partitioner, drifting workloads, and partial entity alignment are natural next steps.

\section*{Declaration on Generative AI}
The authors used Claude (Anthropic) for language editing and structural
refinement. All content was reviewed by the authors, who take responsibility for the
submission. No generative AI was used to produce research findings, experimental results,
or citations. This complies with CEUR's Policy on AI-Assisting Tools.
\bibliography{mybiblography}

\end{document}